\documentclass[]{typhoon}

\usepackage{amsmath,amssymb}
\usepackage{cleveref}
\usepackage{enumitem}
\usepackage{tabularx}
\usepackage{multirow}
\usepackage{graphicx}
\usepackage{xspace}
\usepackage{tikz}
\usepackage{placeins}
\usetikzlibrary{positioning,arrows.meta,decorations.pathreplacing}
\usepackage{fontawesome5}
\colorlet{tyAccent}{typhoonpurple}
\colorlet{tyAccentDark}{typhoonpurple!75!black}
\colorlet{tyAccentFill}{typhoonpurple!10}
\colorlet{tyWarm}{typhoonbg}
\colorlet{tyWarmDark}{typhoonbg!72!black}
\colorlet{tyWarmFill}{typhoonbg!15}
\colorlet{tyNeutral}{black!55}

\tikzset{
  comp/.style={rounded corners=3pt, align=center, text width=3.6cm,
               minimum height=8.5mm, inner sep=3.5pt, draw=black!40, fill=black!6},
  hot/.style={comp, draw=tyAccentDark, fill=tyAccentFill},
  io/.style={align=center, text width=3.6cm, minimum height=7mm,
             inner sep=3pt, draw=black!50, fill=white, rounded corners=1pt},
  side/.style={rounded corners=3pt, align=center, text width=2.7cm,
               minimum height=8.5mm, inner sep=3.5pt},
  opt/.style={side, draw=tyAccentDark, fill=tyAccentFill, dashed},
  cost/.style={font=\tiny, text=black!55, align=center, anchor=north},
  arr/.style={-{Stealth[length=1.8mm]}, black!65, semithick},
  darr/.style={arr, dashed}}

\newcommand{\huggingface}{\raisebox{-1.5pt}{\includegraphics[height=1.05em]{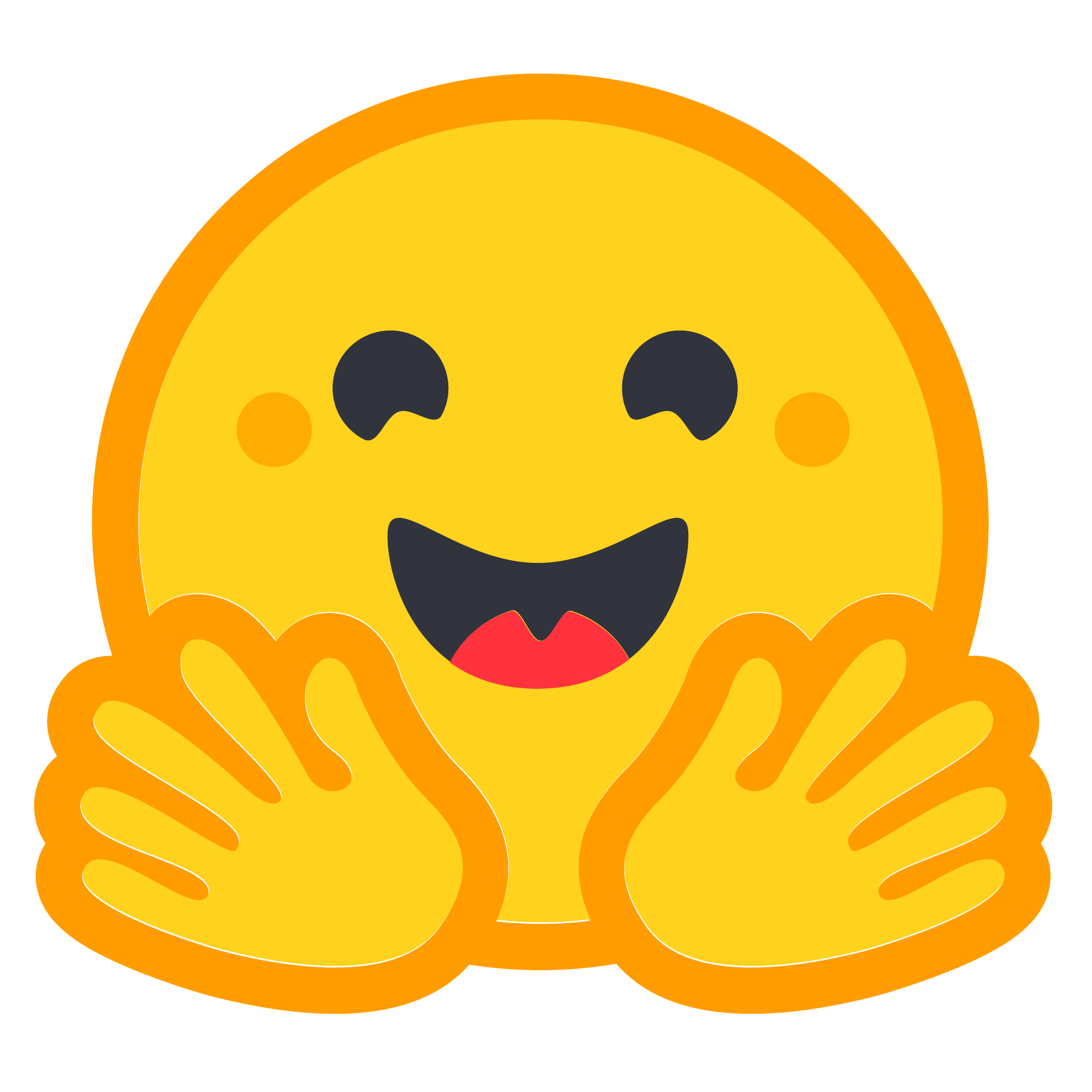}}\xspace}
\newcommand{\github}{\raisebox{-1.5pt}{\includegraphics[height=1.05em]{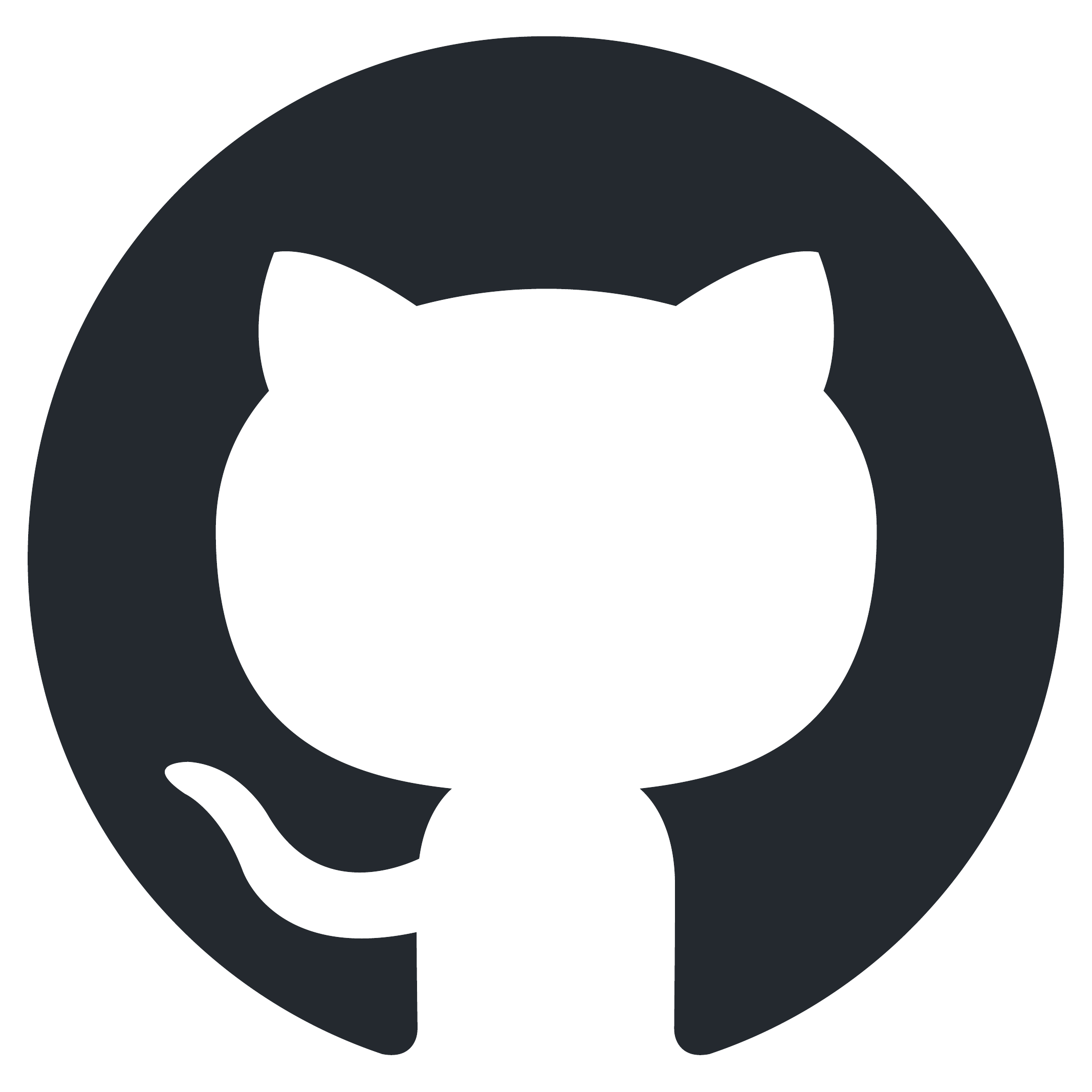}}\xspace}
\newcommand{\typhoonlogo}{\raisebox{-1.5pt}{\includegraphics[height=1.05em]{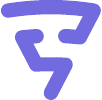}}\xspace}

\setlist[itemize]{leftmargin=*}
\setlist[enumerate]{leftmargin=*}

\newcommand{\bench}[1]{#1}

\newcommand{\pfa}{P(\mathrm{FA})}

\usepackage{xcolor}
\usepackage{booktabs}
\usepackage{array}
\usepackage{colortbl}

\newcommand{\hyb}{Typhoon ASR Edge\xspace}
\newcommand{\hybstream}{Typhoon ASR Edge Streaming\xspace}
\newcommand{\sysname}{Typhoon Keyword Boosting\xspace}
\newcommand{\cab}{\bench{thai-contextasr-bench}\xspace}

\title{Typhoon ASR Edge: A 33M-Parameter Thai Speech Recognition Model with Decode-Time Keyword Boosting}

\author{Warit~Sirichotedumrong, Adisai~Na-Thalang, Shah~Faisal~Wani, Tanawin~Samutsin,
  Sittipong~Sripaisarnmongkol}
\affiliation{Typhoon, SCB DataX}
\abstract{
While open-weight Thai automatic speech recognition (ASR) has achieved strong general accuracy, the most accurate models require GPUs for real-time or high-concurrency use and fail on domain-specific named entities, such as product names, personal names, and rare words. In addition, Thai is written without spaces between words and permits multiple accepted spellings across Thai and Latin scripts, which prevents standard whitespace-based word spotting from being applied directly. We present \hyb, a 33M-parameter FastConformer transducer with an auxiliary connectionist temporal classification (CTC) head, along with \sysname, which spots listed keywords on the CTC posteriors, verifies each candidate under both heads, and rewrites only the verified spans of the transcript at decode time. Despite having only 33M parameters, \hyb provides character error rates (CER) comparable to 115M- to 0.6B-parameter open-weight models on general Thai benchmarks. With a keyword list, \sysname raises entity recall from 32.7\% to 87.8\%, within three percentage points of the strongest proprietary cloud service. Furthermore, a Rust inference engine runs both acoustic inference and keyword boosting with high throughput and low latency on x86-64 CPUs with Intel AMX, with support for other CPU architectures. The model weights, \sysname, and the Rust inference engine are publicly released.
\newline
\huggingface\texttt{Offline Model}: \url{https://huggingface.co/typhoon-ai/typhoon-asr-edge} \newline
\huggingface\texttt{Streaming Model}: \url{https://huggingface.co/typhoon-ai/typhoon-asr-edge-streaming}\newline
\github\texttt{Code \& Evaluation}: \url{https://github.com/warit-s/typhoon-asr-edge}\newline
\github\texttt{Rust Inference Engine}: \url{https://github.com/pongib/typhoon-asr-inference} \newline
\typhoonlogo\texttt{Website}: \url{https://opentyphoon.ai}
}

\begin{document}
\maketitle
\begin{figure}[t]\centering
\begin{subfigure}{\linewidth}\centering
\includegraphics[width=0.668\linewidth]{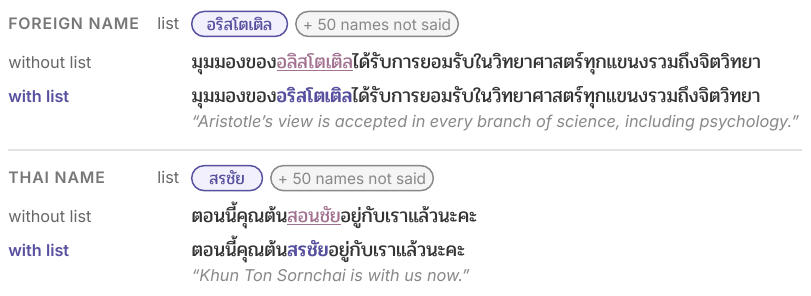}
\caption{Two real \cab utterances, without and with the keyword list.}
\label{fig:hero-examples}
\end{subfigure}

\vspace{3mm}
\begin{subfigure}{\linewidth}\centering
\includegraphics[width=\linewidth]{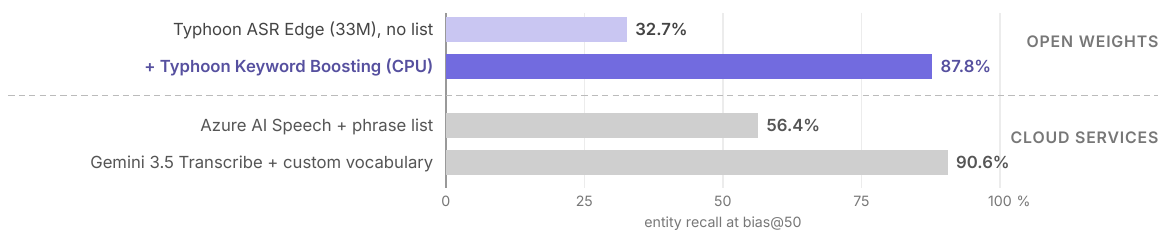}
\caption{Entity recall on \cab at bias@50, where every system receives the same keyword lists.}
\label{fig:hero-recall}
\end{subfigure}
\caption{\textbf{Effect of a keyword list.} Applying \sysname to \hyb on \cab.}
\label{fig:hero}
\end{figure}

\vspace{-20pt}
\section{Introduction}
\label{sec:intro}
Open-weight Thai automatic speech recognition (ASR) has achieved strong general accuracy through Whisper-based models~\citep{radford2023robust,tipaksorn2024PathummaWhisper,aung-etal-2024-thonburian,typhoon_whisper_large}, FastConformer transducers~\citep{typhoonrt,typhoonstreaming,openthai2asrstreaming}, and LLM-based models, such as Qwen3-ASR~\citep{qwen3asr}, Rumble-2-Flash~\citep{rumble2flash}, and Nemotron-3.5-Transcribe~\citep{nemotron35transcribe}. However, ASR systems face limitations when deployed in production environments. First, the most accurate offline and streaming models require GPUs for real-time or high-concurrency use. Second, despite low overall character error rates (CER), existing models fail to transcribe domain-specific named entities, such as product names, personal names, and rare words, which often determine the practical utility of a transcript, especially in Thai where text is written without spaces between words and permits multiple accepted spellings across Thai and Latin scripts for loan words.

To address these limitations, we introduce \hyb, a 33M-parameter FastConformer transducer with an auxiliary connectionist temporal classification (CTC) head, paired with \sysname and a Rust inference engine. Our contributions are as follows:
\begin{itemize}[leftmargin=1.2em, itemsep=2pt]
\item \hyb, an open-weight 33M-parameter Thai ASR model with offline and streaming variants.
\item \sysname, a decode-time keyword boosting method for unsegmented Thai script that enables users to supply a keyword list that steers the transcript, as shown in \Cref{fig:hero}.
\item A Rust inference engine that runs \hyb, \hybstream, and \sysname on CPUs without Python or CUDA.
\end{itemize}

\section{Related Work}
\label{sec:related}
\textbf{Thai ASR.} The open-weight Thai ASR ecosystem consists primarily of Whisper fine-tunes, including Pathumma-Whisper~\citep{tipaksorn2024PathummaWhisper}, Thonburian Whisper~\citep{aung-etal-2024-thonburian}, and Typhoon-Whisper~\citep{typhoon_whisper_large,typhoon_whisper_turbo}. FastConformer transducers, including Typhoon ASR Real-time~\citep{typhoonrt} and Pathumma RNN-T~\citep{pathummaRNNT}, and 0.6B-parameter cache-aware streaming models~\citep{typhoonstreaming,openthai2asrstreaming} have been released as smaller alternatives to Whisper-based models. Multilingual LLM-based ASR models, such as Qwen3-ASR~\citep{qwen3asr}, Rumble-2-Flash~\citep{rumble2flash}, and Nemotron-3.5-Transcribe~\citep{nemotron35transcribe}, have further expanded open-weight multilingual coverage. Apart from Typhoon ASR Real-time and Pathumma RNN-T, all of these models require GPUs for real-time or high-concurrency use.

\textbf{Contextual biasing.} Shallow fusion adds an external language-model score during decoding~\citep{zhao2019shallow,andrusenko2025turbobias,typhoonstreaming}. Alternatively, CTC-based word spotting (CTC-WS)~\citep{andrusenko2024fast,tsai2026streamingctcws} detects listed phrases directly on CTC posteriors and merges them into the greedy transcript without modifying beam search. LLM-based ASR models~\citep{qwen3asr,nemotron35transcribe} have enabled in-prompt contextual biasing by adding keyword lists directly in the text prompt. However, CTC-WS assumes whitespace-delimited word boundaries and has not been validated on unsegmented scripts such as Thai (nor do we evaluate it directly in this work), while LLM-based models require GPUs for deployment and their capacity to scale to realistic keyword lists in Thai has not been investigated.

\textbf{Streaming ASR.} NVIDIA introduced cache-aware streaming FastConformer~\citep{noroozi2024stateful} and the Nemotron streaming architecture built on it~\citep{nemotron35asrstreaming} that bound self-attention and convolution context to emit frames at a fixed look-ahead. In Thai, Typhoon ASR Streaming~\citep{typhoonstreaming} and OpenThai2.0-ASR-Streaming~\citep{openthai2asrstreaming} fine-tune this 0.6B-parameter Nemotron streaming backbone on Thai speech. However, both models rely exclusively on a transducer decoder without an auxiliary CTC head and remain too large for high-concurrency CPU deployment.

\section{Typhoon ASR Edge}
\label{sec:model}
\subsection{Model Architecture}
\label{sec:design}

\hyb is a FastConformer~\citep{rekesh2023fast} hybrid token-and-duration transducer (TDT)~\citep{xu2023tdt} and CTC~\citep{graves2006connectionist} model implemented in NeMo~\citep{kuchaiev2019nemo}, with 16 encoder layers, model dimension $d{=}256$, 4 attention heads, $8{\times}$ subsampling (one frame per 80\,ms), and a CTC loss weight of 0.3.

Since CTC can emit at most one token per encoder frame and Thai speech typically exceeds ten characters per second, a character-level tokenizer would require a higher frame rate. Instead, we use a 2{,}048-piece BPE tokenizer averaging about three characters per piece, which allows the CTC head to operate at 80\,ms frames while keeping the efficiency of $8{\times}$ subsampling.
Unlike other transducer-only Thai ASR, the auxiliary CTC head of \hyb provides a full probability distribution over all vocabulary tokens at every 80\,ms frame to support \sysname, as detailed in \Cref{sec:boosting}. 

\subsection{Training Corpus and Data Preparation}
\label{sec:data}
\Cref{tab:data} lists the 4{,}219.4 hours of training data used to train \hyb.

\begin{table}[!b]\centering\small
\caption{\textbf{Training corpus composition for \hyb}, showing hours retained after filtering and consensus relabeling.}
\label{tab:data}
\setlength{\tabcolsep}{6pt}
\renewcommand{\arraystretch}{1.1}
\begin{tabular}{@{}lrl@{}}
\toprule
\textbf{Source} & \textbf{Hours} & \textbf{License} \\
\midrule
YODAS3 and YODAS2~\citep{li2023yodas,chen2026yodas3} & 3{,}203.2 & CC-BY-3.0 \\
Porjai (Central and Korat)~\citep{suwanbandit23_interspeech} & 716.9 & CC-BY-SA-4.0 \\
Common Voice 17, Thai~\citep{ardila2020common} & 195.0 & CC0 \\
LOTUSDIS, train (mostly close microphones)~\citep{tipaksorn2025lotusdis} & 45.9 & CC-BY-SA-4.0 \\
THAI-SER~\citep{wongpithayadisai2026thaiser} & 30.3 & CC-BY-SA-4.0 \\
Isan conversational speech corpus~\citep{nathalang2025developingopenconversationalspeech} & 19.7 & Apache-2.0 \\
FLEURS, Thai train~\citep{conneau2023fleurs} & 8.4 & CC-BY-4.0 \\
\midrule
\textbf{Total} & \textbf{4{,}219.4} & \\
\bottomrule
\end{tabular}
\end{table}
Since the training corpus comes from multiple sources with different transcription conventions, we relabel the entire corpus under the Typhoon ASR Real-time guideline~\citep{typhoonrt} using teacher consensus across Typhoon Whisper Large V3, Typhoon Whisper Turbo, and Qwen3-ASR-1.7B. Utterances where the teachers disagree are discarded, removing 24.5\% of the source hours.

\subsection{Training Setup}
\label{sec:recipe}
\begin{figure}[!t]\centering
\includegraphics[width=\linewidth]{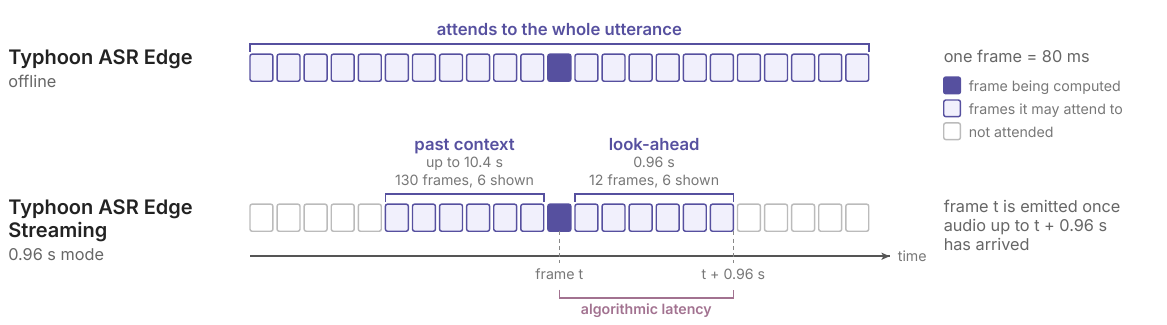}
\caption{\textbf{Receptive field of one encoder frame} (80\,ms per frame). The dark box denotes the frame being computed and the shaded boxes denote the frames it can attend to. In offline mode, each frame attends to the entire utterance. In the default streaming mode, attention is restricted to 130 past frames and 12 future frames (drawn as 6 and 6 for illustration). Past frames add no latency, while the 12 future frames determine the 0.96\,s look-ahead. The same checkpoint also supports a [70,\,6] window (0.48\,s look-ahead) and full-context inference.}
\label{fig:context}
\end{figure}

We train \hyb from scratch on the 4{,}219.4-hour corpus as shown in \Cref{tab:data} using full-utterance attention and non-causal convolutions. \hybstream is then initialized from \hyb and fine-tuned with causal convolutions and a bounded temporal receptive field, as illustrated in \Cref{fig:context}. To support both streaming look-ahead windows (0.96\,s and 0.48\,s) and offline inference within a single checkpoint, we apply dual-mode training~\citep{yu2021dualmode}, encoding each batch in both full-context and streaming modes with a KL divergence loss between their CTC posteriors. During streaming fine-tuning, we also concatenate utterances up to 40\,s to prevent the model from collapsing on long audio sequences. Complete training hyperparameters for both models are listed in Appendix~\ref{app:hparams}.

\section{\sysname}
\begin{figure}[!b]\centering
\includegraphics[width=\linewidth]{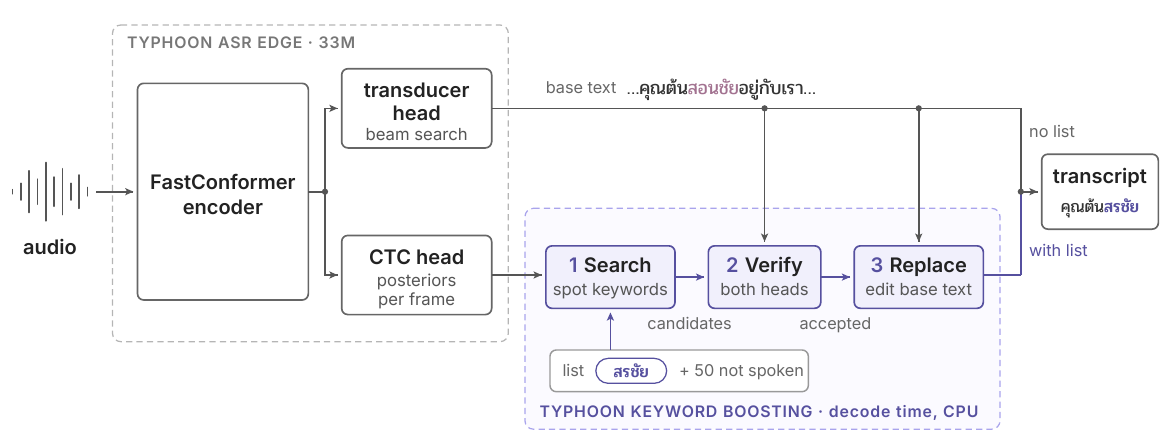}
\caption{\textbf{Overview of \sysname.} The encoder processes the input audio and feeds both the transducer and CTC heads. The transducer head produces the base text for keyword boosting. When a keyword list is given, (1) the search stage spots candidate frame spans on the CTC posteriors, (2) the verification stage scores each candidate under both heads, and (3) the replacement stage writes the accepted keywords into the base text.}
\label{fig:system}
\end{figure}

\label{sec:boosting}\label{sec:system}
In Thai ASR, personal names, product names, and rare domain terms often sound identical to common words or admit multiple valid spellings across Thai and Latin scripts, so the intended spelling cannot be inferred from audio alone. \sysname addresses this by allowing users to supply a keyword list that is applied at decode time in three CPU stages without model retraining, as illustrated in \Cref{fig:system}. First, the encoder processes the input audio and feeds both the transducer and CTC heads.

The transducer head produces the base text for keyword boosting (the best of eight beam hypotheses rescored with the CTC head). Because beam search prunes low-probability token sequences, unfamiliar named entities are often transcribed as common words, such as \textthai{สรชัย} transcribed as \textthai{สอนชัย} in \Cref{fig:hero}. In contrast, the CTC head outputs a log-posterior matrix $\mathbf{L}\in\mathbb{R}^{T\times 2049}$ over all vocabulary pieces at every 80\,ms frame, retaining probability mass for the pieces of \textthai{สรชัย} at the corresponding frames. Given the base text and the CTC posteriors, \sysname applies the keyword list in three stages: (1) \textbf{Search} spots candidate frame spans for each listed keyword on the CTC posteriors, (2) \textbf{Verification} scores each candidate under both the transducer and CTC heads to filter out false positives, and (3) \textbf{Replacement} maps accepted frame spans onto the base text and writes the listed spellings.

\begin{figure}[!b]\centering
\includegraphics[width=\linewidth]{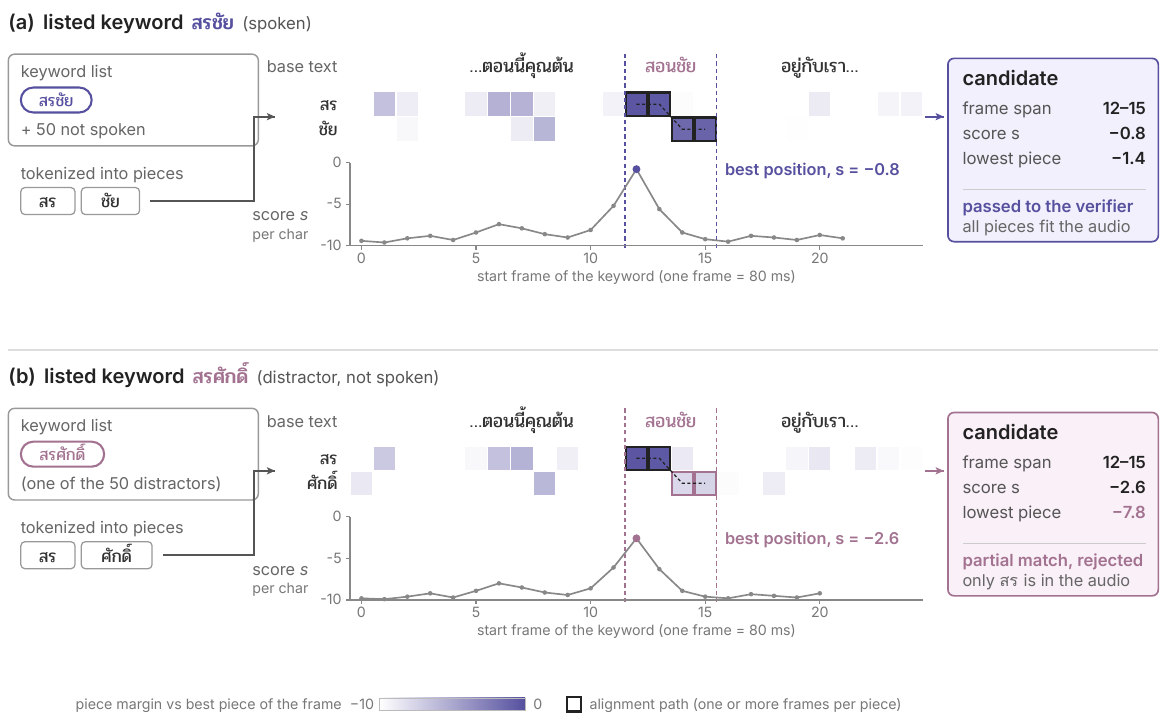}
\caption{\textbf{Search stage} on the second utterance of \Cref{fig:hero} (illustrative scores). A listed keyword is tokenized into vocabulary pieces and slid across the CTC frame posteriors, where each piece is scored relative to the highest-scoring token at its frame. (a) \textthai{สรชัย} matches the frames where the base text decoded \textthai{สอนชัย}, and the search outputs its frame span, average score per character, and lowest single-piece score. (b) The unspoken distractor \textthai{สรศักดิ์} matches the initial syllable and achieves a reasonable average score, but its final piece scores poorly against the audio, allowing the lowest single-piece score to flag and reject partial matches during candidate filtering and verification.}
\label{fig:search}
\end{figure}

\subsection{Search}
As illustrated in \Cref{fig:search}, the search stage scans the CTC posteriors to find the frame span that best matches each listed keyword. A keyword is tokenized into vocabulary pieces and aligned against the frame posteriors, allowing short pauses between pieces and at most one skipped character. At each frame, a piece is scored relative to the most probable piece. Because this search runs on the full posterior distribution rather than the decoded text, it can recover keywords missed by the transducer, such as spotting \textthai{สรชัย} at the frames decoded as \textthai{สอนชัย}. Each candidate match passes three values to the verifier: its frame span, its average score per character, and its lowest single-piece score, which helps reject partial matches where only part of the keyword fits the audio, such as the distractor \textthai{สรศักดิ์} in \Cref{fig:search}b.

To support English and other Latin-script keywords, which are typically pronounced under Thai phonology, we expand each Latin entry into Thai phonetic spellings using a character-level transliterator and two rule-based mappings (covering letter-by-letter acronyms and English-to-Thai grapheme rules). The search matches these Thai spellings against the CTC posteriors, while preserving the original listed spelling for the final transcript.

\subsection{Verification}
\begin{figure}[!htb]\centering
\includegraphics[width=\linewidth]{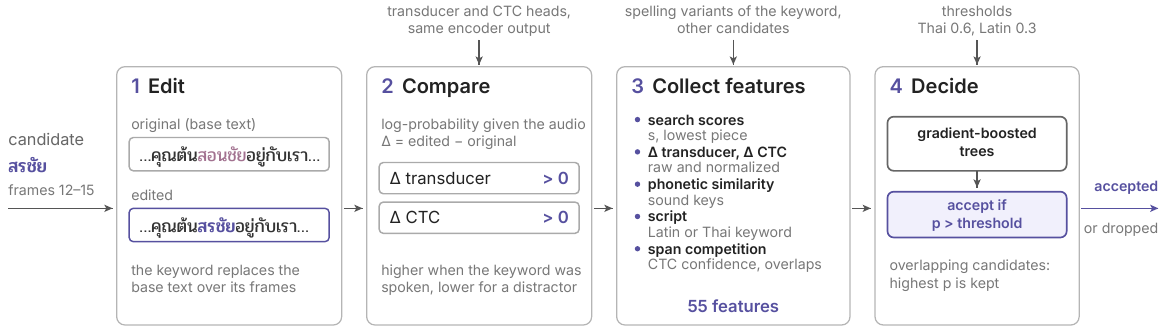}
\caption{\textbf{Verification stage.} The candidate keyword is substituted into the base text, the original and edited transcripts are scored against the audio under both output heads, additional context features are extracted, and a gradient-boosted tree classifier decides whether to accept the candidate.}
\label{fig:verify}
\end{figure}

Since the search stage aims to maximize recall, the verification stage checks whether each candidate matches the audio to avoid false positives. The verifier substitutes the candidate keyword into the base text and evaluates whether the edited transcript better matches the audio in four steps, as shown in \Cref{fig:verify}.
\begin{enumerate}[label=Step \arabic*:, leftmargin=1.6cm, labelsep=0.4em, itemsep=2pt, align=left]
\item \textbf{Edit.} The candidate keyword replaces the base text over its matched frame span.
\item \textbf{Compare against the audio.} Both the transducer and CTC heads compute the log-probability of the original and edited transcripts given the audio, and take the score difference. The edited transcript scores higher than the original when the keyword was spoken, and lower when it is an unspoken distractor.
\item \textbf{Collect additional features.} Probability differences alone can fail for Latin keywords. For example, replacing \textthai{แอลวีเอ็มเอช} with \texttt{LVMH} lowers the transducer probability even though the pronunciation is identical, because the transducer favors Thai script. The verifier therefore adds three groups of features: (1) phonetic similarity, measured by reducing both the replaced text and the keyword to consonant-and-vowel keys that ignore tone marks and silent letters and merge consonants of the same sound; (2) script indicators that specify whether the keyword is written in Latin script, where a lower transducer probability is expected; and (3) span competition, which records the CTC confidence within the span and how many other candidates overlap the same frames. Combined with the search scores and probability differences, these yield 55 features in total, as listed in Appendix~\ref{app:features}.
\item \textbf{Decide.} A gradient-boosted decision tree classifier~\citep{friedman2001gbm,pedregosa2011scikit} takes the 55 features and predicts the probability that each candidate is valid. Candidates scoring above 0.6 for Thai or 0.3 for Latin keywords are accepted in descending order of probability, while lower-scoring candidates on overlapping frames are discarded. The classifier and its acceptance thresholds are fitted on an internal banking-domain biasing benchmark (synthetic speech with human-verified transcripts and entity annotations) and are not tuned on \cab.
\end{enumerate}

\subsection{Replacement}
\begin{figure}[!t]\centering
\includegraphics[width=\linewidth]{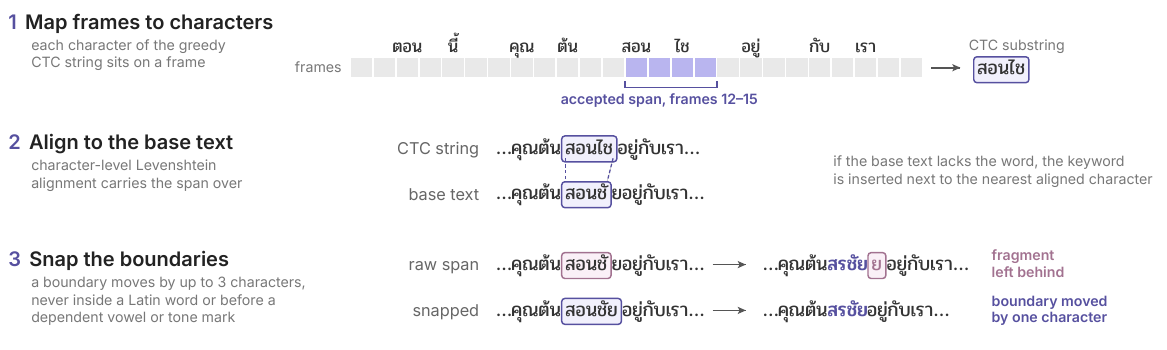}
\caption{\textbf{Replacement stage.} The accepted frame span is mapped to characters via the greedy CTC string (Step 1), aligned to the base text using character-level alignment (Step 2), and snapped to valid word and syllable boundaries before substituting the listed keyword spelling (Step 3).}
\label{fig:replace}
\end{figure}

An accepted candidate specifies a frame span, whereas the base text is a character string without frame timestamps. The replacement stage converts the frame span into a character span in the base text in three steps, as shown in \Cref{fig:replace}.
\begin{enumerate}[label=Step \arabic*:, leftmargin=1.6cm, labelsep=0.4em, itemsep=2pt, align=left]
\item \textbf{Map frames to characters.} Because each character in the greedy CTC output is associated with a frame index, the accepted frame span maps directly to a CTC substring, such as \textthai{สอนไช} for frames 12--15.
\item \textbf{Align to the base text.} Character-level Levenshtein alignment between the greedy CTC string and the base text maps the CTC substring to a corresponding span in the base text. If the spoken word is completely missing from the base text, the keyword is inserted next to the nearest aligned character.
\item \textbf{Snap the boundaries.} The aligned span may end inside a word or before a dependent vowel or tone mark, which would leave stray characters behind (such as the trailing \textthai{ย} in \Cref{fig:replace}, ``P\textthai{โพเมโล่}'', or ``SiSilicon Valleyey''). To prevent this, each span boundary can shift by up to three characters toward the position that best matches the keyword phonetically, while avoiding splits inside a Latin word or immediately before a Thai dependent vowel or tone mark. Finally, the listed keyword spelling replaces the adjusted span.
\end{enumerate}

\section{Rust Inference Engine}
\label{sec:engine}
\begin{figure}[!t]\centering
\includegraphics[width=\linewidth]{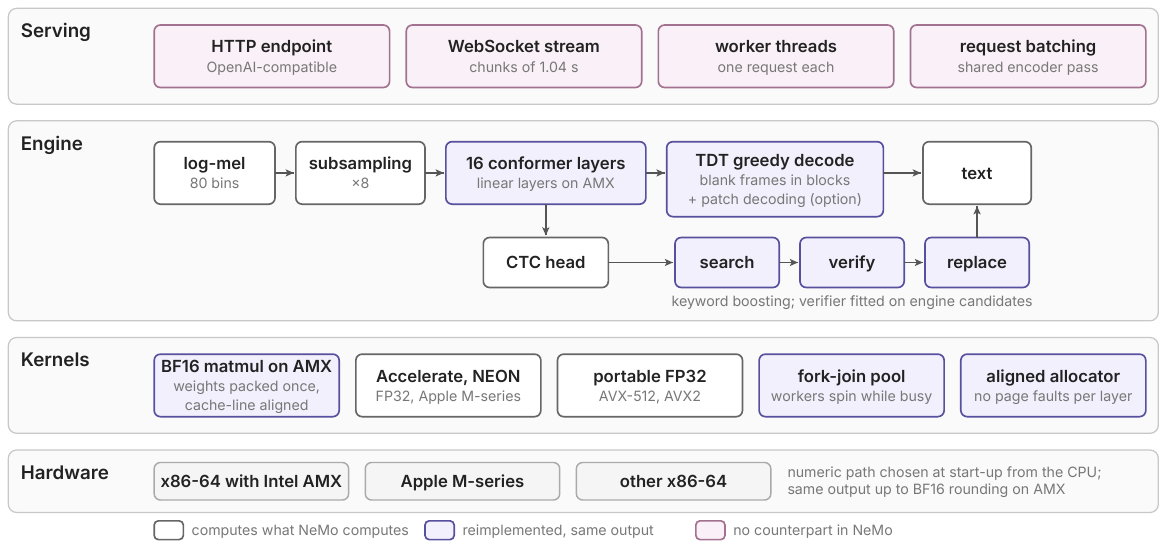}
\caption{\textbf{Architecture of the Rust inference engine.} Each request enters through the serving interface and runs on a worker thread using architecture-specific CPU kernels, while keyword boosting reuses the CTC posteriors from the same encoder pass. White blocks match NeMo's standard pipeline, purple blocks are optimized reimplementations that produce identical outputs, and warm-shaded blocks denote components unique to our engine.}
\label{fig:engine}
\end{figure}

We release a CPU inference engine written in Rust that runs \hyb in offline and streaming modes without Python or CUDA, and reproduces the transcripts of NeMo.\footnote{\url{https://github.com/pongib/typhoon-asr-inference}} As illustrated in \Cref{fig:engine}, the architecture consists of four layers: a serving layer, a single-utterance execution engine, low-level compute kernels, and the target CPU hardware. It follows the exact network computation and greedy decoding of NeMo while optimizing weight storage, matrix multiplication, and thread scheduling.

\subsection{Kernels}
The low-level compute layer accelerates matrix operations and minimizes memory and scheduling overhead:
\begin{itemize}
\item \textbf{Matrix Multiplication.} The engine dispatches matrix multiplication according to the detected CPU architecture, using bfloat16 (BF16) with Intel Advanced Matrix Extensions (AMX) on supported x86-64 processors, Apple Accelerate and Arm NEON on Apple M-series processors, and 32-bit floating point (FP32) on other processors.
\item \textbf{Weight Layout.} When loading a checkpoint, the engine packs weights into the AMX tile format with cache-line alignment, which is $6.5{\times}$ faster to load than unaligned tiles.
\item \textbf{Weight Reuse.} Since CPU inference is bottlenecked by memory bandwidth rather than compute, the engine computes each feed-forward block without storing intermediate matrices in memory, and loads the model weights only once per batch of requests.
\item \textbf{Thread Pool.} Transcribing an utterance involves around 200 parallel operations, each taking only tens of microseconds. To avoid the cost of waking threads through the operating system, worker threads spin while a transcription is in progress instead of sleeping between operations.
\end{itemize}

\subsection{Engine}
On top of the compute kernels, the single-utterance execution engine runs the acoustic model, decoding, and keyword boosting:
\begin{itemize}
\item \textbf{Greedy Decoding.} The engine implements NeMo's greedy transducer decoding. Instead of evaluating the decoder frame by frame, it scores consecutive blank frames together in a single matrix multiplication and updates the decoder state only when a non-blank token is emitted.
\item \textbf{Patch Decoding.} We also introduce an optional decoding mode beyond NeMo's greedy decoding. At frames where the top two token or duration candidates have similar scores, the decoder follows the second candidate until its path rejoins the greedy path. If the alternative span scores higher under both the transducer and CTC heads, it replaces the greedy span. The results of \Cref{sec:plain,sec:biasing} are decoded with NeMo and therefore without patch decoding, while \Cref{tab:engine} benchmarks throughput with patch decoding enabled.
\item \textbf{Keyword Boosting.} The engine implements the three stages of \sysname described in \Cref{sec:boosting}, reusing the CTC posteriors from the same encoder pass. Its verifier is trained on candidate outputs from the Rust engine.
\item \textbf{Streaming.} Across consecutive audio chunks, the engine retains per-layer encoder caches and carries over the decoder state without revising previously emitted text. When keyword boosting is combined with streaming, replacement is applied at the end of an utterance or endpointed segment rather than revising already emitted chunks.
\end{itemize}

\subsection{Serving}
The engine includes a built-in server to handle concurrent client requests over standard network protocols:
\begin{itemize}
\item \textbf{Worker Threads.} Each transcription request executes on a single worker thread. When multiple requests accumulate in the queue, an available worker batches them together through the encoder so that the model weights are read only once, while returning results identical to running each request individually.
\item \textbf{Server Process.} A single server process loads both the offline and streaming checkpoints, exposes an OpenAI-compatible HTTP endpoint alongside a WebSocket streaming endpoint, and uses 0.4\,GB of memory for \hyb.
\end{itemize}

\section{Experiments}
\label{sec:experiments}
\subsection{Setup}
\label{sec:protocol}

\subsubsection{Benchmarks}
We evaluate our models on four benchmarks covering general transcription, keyword boosting, and meeting speech:
\begin{itemize}
\item \textbf{General Thai ASR:} \bench{gigaspeech2-typhoon}~\citep{typhoon_gigaspeech_test} (1{,}000 human-verified YouTube segments of 2--10\,s) and \bench{TVSpeech}~\citep{tvspeech_dataset} (570 talk-show segments of up to 30\,s).
\item \textbf{Keyword Boosting:} \cab~\citep{contextasrbench} (2{,}062 conversational utterances, roughly 33\% real speech and the remainder voice-cloned synthetic speech, with official keyword lists and evaluation scripts). The even-indexed half of \cab was among the human-verified utterances used to build the training labels in \Cref{sec:data}, while the odd-indexed half is untouched by any training or tuning. We report results on the full set.
\item \textbf{Meeting Speech:} LOTUSDIS test set~\citep{tipaksorn2025lotusdis} (5{,}516 meeting turns evaluated across three close-microphone channels: lapel, tabletop conference microphone, and speakerphone).
\end{itemize}

\subsubsection{Metrics and List Conditions}
\label{sec:conditions}
For general ASR, we report macro-averaged CER after text normalization, which strips whitespace and normalizes digits versus Thai number words, Latin versus Thai loanword spellings, and colloquial particles, such as \textthai{มั้ย} and \textthai{ไหม}. Per-utterance CER is capped at 100\% so that a single repetition loop cannot dominate the macro average. Applying consistent normalization across all systems separates acoustic recognition errors from formatting differences across Thai datasets and model conventions. For decoding, \Cref{tab:cer,tab:stream} use greedy decoding for \hyb and \hybstream, whereas \Cref{tab:biasing,tab:routes} run the Python/NeMo pipeline with the beam-8 CTC-rescored base text described in \Cref{sec:boosting}.

For keyword boosting on \cab, we evaluate three list conditions per utterance:
\begin{itemize}
\item \textbf{none:} no keyword list is provided.
\item \textbf{bias@50:} spoken target entities plus 50 unspoken distractors.
\item \textbf{bias@500:} spoken target entities plus 500 unspoken distractors.
\end{itemize}
In addition to CER, we report \textbf{entity recall} (the percentage of spoken target entities transcribed in the listed spelling) and the \textbf{false-alarm rate ($\pfa$)} (the percentage of unspoken distractors falsely inserted per distractor under the corresponding list condition).

\subsubsection{Baselines}
Open-weight baselines are decoded using the settings from their model cards unless noted otherwise:
\begin{itemize}
\item \textbf{Offline ASR:} Whisper Large V3 and its Thai fine-tunes~\citep{tipaksorn2024PathummaWhisper,aung-etal-2024-thonburian,typhoon_whisper_large}, Qwen3-ASR-1.7B~\citep{qwen3asr}, Rumble-2-Flash~\citep{rumble2flash}, Typhoon ASR Real-time~\citep{typhoonrt}, Azure AI Speech, Gemini 3.5 Transcribe, and Nemotron-3.5-Transcribe~\citep{nemotron35transcribe} served via vLLM at temperature 0.
\item \textbf{Streaming ASR:} Typhoon ASR Streaming~\citep{typhoonstreaming} (0.6B) and OpenThai2.0-ASR-Streaming~\citep{openthai2asrstreaming} (0.64B, using its documented blank offset), both evaluated at 0.48\,s and 1.0\,s look-ahead.
\item \textbf{Keyword Boosting:} We compare against prompt biasing on Nemotron-3.5-Transcribe, Qwen3-ASR-1.7B (both zero-shot and with a LoRA adapter fine-tuned on an internal 10{,}846-hour Thai corpus using keyword lists and 5\% negative batches), and Rumble-2-Flash (passing the keyword list through an undocumented prompt input, as its model card does not document contextual biasing or Thai benchmarks). We also evaluate Azure AI Speech (\texttt{th-TH}) with phrase lists and Gemini 3.5 Transcribe with custom vocabulary, queried in October and November 2026 with identical lists. Finally, to test whether \sysname transfers beyond \hyb, we run it on Typhoon ASR Real-time by adding a 4-layer CTC head on top of its frozen encoder.
\end{itemize}
\subsection{Results}
\subsubsection{General Transcription}
\label{sec:plain}
\Cref{tab:cer} shows offline transcription results across the three benchmarks using normalized macro CER. Typhoon Whisper Large V3 achieves the lowest CER on \bench{gigaspeech2-typhoon} and \bench{TVSpeech}, while Gemini 3.5 Transcribe and Qwen3-ASR-1.7B lead on the close-microphone channels of LOTUSDIS. Among models under 1B parameters, \hyb and the full-context mode of \hybstream outperform Rumble-2-Flash (0.5B) and Thonburian Whisper Medium (769M) on both general benchmarks, and perform closely to the 115M-parameter Typhoon ASR Real-time on \bench{TVSpeech} (7.38\% and 7.51\% compared to 7.23\%) and LOTUSDIS (29.31--31.58\% compared to 29.32\%), while trailing it by 0.94--1.13 points on \bench{gigaspeech2-typhoon}.

\begin{table}[!b]\centering\footnotesize
\caption{Offline ASR results without a keyword list (normalized macro CER in \%, following \Cref{sec:conditions}). LOTUSDIS is evaluated on the three close microphones of the test set (lapel, conference microphone, and speakerphone, 5{,}516 meeting turns each, 16{,}548 recordings in total), reporting the mean of the per-microphone macro CERs. Nemotron-3.5-Transcribe is evaluated using its Thai context prompt paired with a fixed list of 50 unrelated Thai words to anchor the output language, though it still outputs English translations on 45\% of \bench{TVSpeech} clips.}
\label{tab:cer}
\setlength{\tabcolsep}{6pt}
\renewcommand{\arraystretch}{1.1}
\begin{tabular}{@{}lrrrr@{}}
\toprule
\textbf{Model} & \textbf{Params} & \bench{gigaspeech2-typhoon} & \bench{TVSpeech} & LOTUSDIS (close mics) \\
\midrule
Gemini 3.5 Transcribe & -- & 4.95 & 5.06 & \textbf{23.24} \\
Azure AI Speech & -- & 5.48 & 7.10 & 25.15 \\
\midrule
Typhoon Whisper Large V3 & 1.55B & \textbf{4.51} & \textbf{4.74} & 26.41 \\
Pathumma Whisper Large V3 & 1.55B & 4.77 & 7.03 & 24.43 \\
Thonburian Whisper Large V3 & 1.55B & 6.43 & 11.22 & 25.26 \\
Whisper Large V3 & 1.55B & 9.41 & 14.86 & 31.98 \\
Thonburian Whisper Medium & 769M & 7.80 & 11.49 & 26.74 \\
Qwen3-ASR-1.7B & 1.7B & 5.35 & 7.70 & 24.09 \\
Nemotron-3.5-Transcribe & 34B & 10.71 & 44.26 & 36.71 \\
Rumble-2-Flash & 0.5B & 7.29 & 9.75 & 48.79 \\
Typhoon ASR Real-time & 115M & 5.50 & 7.23 & 29.32 \\
\rowcolor{typhoonpurple!12}
\textbf{\hyb} & \textbf{33M} & 6.63 & 7.38 & 31.58 \\
\rowcolor{typhoonpurple!12}
\textbf{\hybstream}, full-context mode & \textbf{33M} & 6.44 & 7.51 & 29.31 \\
\bottomrule
\end{tabular}
\end{table}

\Cref{tab:stream} summarizes streaming transcription results. \hybstream achieves the lowest CER on \bench{TVSpeech} at 0.96\,s look-ahead (8.72\%) and stays within 1.4 CER of the offline \hyb model across both general benchmarks. On \bench{gigaspeech2-typhoon}, Typhoon ASR Streaming achieves the lowest CER (6.63\%), and on LOTUSDIS (close mics), OpenThai2.0-ASR-Streaming achieves the lowest CER (21.58\%).

\begin{table}[!t]\centering\footnotesize
\caption{Streaming ASR results without a keyword list (normalized macro CER in \%, following \Cref{sec:conditions}). Look-ahead denotes the algorithmic latency of the encoder, and the full-context mode of \hybstream is reported in \Cref{tab:cer}. LOTUSDIS is evaluated on the three close microphones as in \Cref{tab:cer}.}
\label{tab:stream}
\setlength{\tabcolsep}{6pt}
\renewcommand{\arraystretch}{1.1}
\begin{tabular}{@{}lrlrrr@{}}
\toprule
\textbf{Model} & \textbf{Params} & \textbf{Look-ahead} & \bench{gigaspeech2-typhoon} & \bench{TVSpeech} & LOTUSDIS (close mics) \\
\midrule
\rowcolor{typhoonpurple!12}
 & & 0.48\,s & 8.61 & 9.84 & 32.12 \\
\rowcolor{typhoonpurple!12}
\multirow{-2}{*}{\textbf{\hybstream}} & \multirow{-2}{*}{33M} & 0.96\,s & 7.64 & \textbf{8.72} & 30.42 \\
\midrule
\multirow{2}{*}{Typhoon ASR Streaming} & \multirow{2}{*}{0.6B} & 0.48\,s & 6.66 & 9.97 & 29.69 \\
 & & 1.0\,s & \textbf{6.63} & 9.88 & 29.50 \\
\midrule
\multirow{2}{*}{OpenThai2.0-ASR-Streaming} & \multirow{2}{*}{0.64B} & 0.48\,s & 6.67 & 10.14 & 21.81 \\
 & & 1.0\,s & 6.78 & 9.75 & \textbf{21.58} \\
\bottomrule
\end{tabular}
\end{table}

\subsubsection{Keyword Boosting}
\label{sec:biasing}
\Cref{tab:biasing} presents keyword boosting results on \cab. \sysname raises the entity recall of \hyb at bias@50 from 32.7\% to 87.8\% and reduces normalized CER by 1.80 points (from 6.65\% to 4.85\%) while keeping false alarms at 0.03\% per distractor (0.015 false alarms per utterance). Applying \sysname to \hybstream and to Typhoon ASR Real-time with an added CTC head yields similar recall and CER improvements, suggesting that the pipeline transfers across FastConformer models.

\begin{table}[!b]\centering\small
\caption{Keyword boosting results on \cab (full set, in \%). \emph{CER} applies the normalization protocol of \Cref{sec:conditions} to the benchmark reference text, while \emph{Recall} and $\pfa$ are computed by the official benchmark scorer. The 95\% confidence intervals of the scorer are approximately $\pm$0.35 CER and $\pm$1.3 recall points. $\pfa$ is reported in \% per distractor at bias@50.}
\label{tab:biasing}
\setlength{\tabcolsep}{6pt}
\renewcommand{\arraystretch}{1.1}
\begin{tabular}{@{}lrrrrr@{}}
\toprule
 & \multicolumn{2}{c}{\textbf{none}} & \multicolumn{3}{c}{\textbf{bias@50}} \\
\cmidrule(lr){2-3}\cmidrule(lr){4-6}
\textbf{System} & CER$\downarrow$ & Recall$\uparrow$ & CER$\downarrow$ & Recall$\uparrow$ & $\pfa\downarrow$ \\
\midrule
\rowcolor{typhoonpurple!12}
\hyb & 6.65 & 32.7 & \textbf{4.85} & \textbf{87.8} & \textbf{0.03} \\
\hybstream, 0.96\,s & 7.07 & 31.6 & 5.29 & 87.8 & 0.05 \\
Typhoon ASR Real-time + CTC head & 6.59 & 32.0 & \textbf{4.85} & 86.4 & 0.07 \\
\bottomrule
\end{tabular}
\end{table}

\begin{table}[!htb]\centering\scriptsize
\caption{Comparison of three keyword-biasing approaches on the full \cab set (all values in \%). \emph{CER} applies the normalization protocol of \Cref{sec:conditions} at bias@50, while \emph{Recall} and $\pfa$ are computed by the official benchmark scorer. $\pfa$ is the percentage of unspoken distractors that appear in the transcript under the list of that condition. Prompt-based systems receive the list as context terms in the prompt (Rumble-2-Flash is run with spelled-out numbers), whereas commercial services use their native vocabulary inputs.}
\label{tab:routes}
\setlength{\tabcolsep}{4pt}
\renewcommand{\arraystretch}{1.1}
\begin{tabular}{@{}llrrrrrrr@{}}
\toprule
 & & & \textbf{none} & \multicolumn{3}{c}{\textbf{bias@50}} & \multicolumn{2}{c}{\textbf{bias@500}} \\
\cmidrule(lr){4-4}\cmidrule(lr){5-7}\cmidrule(lr){8-9}
\textbf{Route} & \textbf{Model} & \textbf{Params} & Recall$\uparrow$ & CER$\downarrow$ & Recall$\uparrow$ & $\pfa\downarrow$ & Recall$\uparrow$ & $\pfa\downarrow$ \\
\midrule
prompt & Rumble-2-Flash & 0.5B & 31.3 & 8.23 & 32.4 & 0.02 & 16.6 & 0.02 \\
prompt & Qwen3-ASR-1.7B & 1.7B & 39.1 & 6.18 & 52.3 & 0.43 & 41.1 & 0.01 \\
prompt & Qwen3-ASR-1.7B, fine-tuned with keyword lists & 1.7B & 44.1 & 4.32 & 64.0 & \textbf{0.01} & 48.3 & \textbf{0.00} \\
prompt & Nemotron-3.5-Transcribe & 34B & 43.6 & 5.86 & 75.4 & 0.15 & 37.0$^\ast$ & 19.4 \\
service & Azure AI Speech + phrase list & -- & 45.3 & 4.65 & 56.4 & \textbf{0.01} & 55.4 & 0.01 \\
service & Gemini 3.5 Transcribe + custom vocab.\ & -- & \textbf{65.7} & \textbf{2.76} & \textbf{90.6} & \textbf{0.01} & \textbf{82.1} & 0.01 \\
\rowcolor{typhoonpurple!12}
decode-time & \hyb + \sysname & 33M & 32.7 & 4.85 & 87.8 & 0.03 & 80.2 & 0.01 \\
\bottomrule
\end{tabular}\\[2pt]
{\scriptsize $^\ast$ Model copies list entries into the output instead of transcribing the audio (34\% of hypotheses exceed three times the reference length, with CER above 1{,}000\%).}
\end{table}

\Cref{tab:routes} compares decode-time keyword boosting against prompt-based biasing and commercial cloud APIs. Zero-shot prompt biasing degrades on distractor-heavy lists. While Nemotron-3.5-Transcribe improves recall at bias@50, zero-shot prompt models drop sharply at bias@500, where Rumble-2-Flash falls into repetition loops (26.8\% of utterances) and Nemotron-3.5-Transcribe copies unspoken list entries into the transcript (19.4\% $\pfa$). Fine-tuning Qwen3-ASR-1.7B with keyword lists improves recall at bias@50, though performance still drops at bias@500. Overall, \hyb with \sysname is the only open-weight system that maintains high entity recall and low false-alarm rates at both bias@50 (0.03\% per distractor) and bias@500 (0.01\% per distractor, or 0.05 false alarms per utterance), coming within 2.8 recall points of Gemini 3.5 Transcribe at bias@50 (though with a higher CER). These results show that spotting candidates on CTC posteriors and verifying each edit under both output heads scales far more reliably to hundreds of distractors than conditioning autoregressive decoders on prompt lists.

\subsubsection{Inference Engine Parity and Performance}
\label{sec:engine-eval}
We evaluate the Rust inference engine against unoptimized PyTorch FP32 NeMo on a virtual machine with four Intel Xeon Platinum 8481C (Sapphire Rapids) cores and eight hardware threads using Intel AMX BF16 (the engine also supports Apple Silicon and generic FP32 CPUs, which are not benchmarked here). \Cref{tab:engine-cer} compares transcription accuracy and output agreement on \bench{gigaspeech2-typhoon} and \bench{TVSpeech}, confirming that the engine matches NeMo's transcripts exactly in FP32 and closely in BF16 while optional patch decoding further lowers CER. \Cref{tab:engine} reports single-request latency, server throughput, and streaming capacity, showing substantial speedups over PyTorch FP32 NeMo on offline workloads alongside high streaming capacity.

\begin{table}[!htb]\centering\small
\caption{Comparison between NeMo and the Rust inference engine for \hyb on \bench{gigaspeech2-typhoon} / \bench{TVSpeech}. ``Identical'' denotes the percentage of transcripts matching FP32 NeMo character for character. CER follows the engine repository's evaluation protocol (micro CER after Unicode NFC normalization, Thai \emph{sara am} normalization, lowercasing, and removing whitespace and punctuation), which differs from the macro CER in \Cref{tab:cer}.}
\label{tab:engine-cer}
\setlength{\tabcolsep}{5pt}
\renewcommand{\arraystretch}{1.1}
\begin{tabular}{@{}lcc@{}}
\toprule
\textbf{Decode} & \textbf{CER (\%)} & \textbf{Identical (\%)} \\
\midrule
NeMo, greedy, FP32 & 6.63 / 8.41 & reference \\
Engine, greedy, FP32 & 6.63 / 8.41 & 100.0 / 100.0 \\
Engine, greedy, BF16 & 6.61 / 8.40 & 99.3 / 92.8 \\
Engine, patch decoding, BF16 & 6.29 / 8.08 & N/A \\
\bottomrule
\end{tabular}
\end{table}

\begin{table}[!htb]\centering\small
\caption{Inference latency and throughput on a 4-core Intel Xeon Platinum 8481C instance (8 hardware threads, AMX BF16, patch decoding enabled). Single-request latency is averaged over \bench{gigaspeech2-typhoon} (p99 in parentheses). Real-time capacity is computed as audio-seconds processed per wall-second across 16 concurrent streams. N/A: Typhoon ASR Real-time has no streaming mode.}
\label{tab:engine}
\setlength{\tabcolsep}{5pt}
\renewcommand{\arraystretch}{1.1}
\begin{tabular}{@{}lrr@{}}
\toprule
 & \textbf{\hyb} & \textbf{Typhoon ASR Real-time} \\
\midrule
single request, 1 thread & 15.8\,ms (34) & 33.6\,ms (66) \\
single request, 8 threads & 6.5\,ms (13) & 12.9\,ms (25) \\
long form, 23.7\,s utterance, 8 threads & 40.4\,ms & 80.1\,ms \\
NeMo (PyTorch FP32), 8 threads & 56.7\,ms & 102.2\,ms \\
\midrule
server, 8 workers, 8 clients & 17{,}019 req/min & 7{,}389 req/min \\
server, 4 workers, saturated (batch 8) & 18{,}671 req/min & 10{,}106 req/min \\
\midrule
streaming, compute per chunk & 6.9\,ms & N/A \\
real-time capacity (audio-s / wall-s) & 651 & N/A \\
\bottomrule
\end{tabular}
\end{table}

\section{Limitations}
\label{sec:limits}
\textbf{\sysname and verifier training data.} The verification classifier is trained on 496 positive candidates from synthetic banking-domain speech. Transferring the verifier and its acceptance thresholds directly to other domains reduces entity recall unless the threshold is recalibrated on in-domain data.

\textbf{Evaluation scope.} This paper does not include component-level ablations of the three \sysname stages or its 55 verifier features, nor does it directly benchmark CTC-WS, shallow fusion, or Pathumma RNN-T under the same protocol. In addition, we do not measure the standalone runtime latency of keyword boosting separately from acoustic inference, and we do not evaluate code-switching or regional Thai dialects.

\textbf{Far-field speech and conventions.} Our evaluation focuses on close-microphone speech and excludes the distant far-field channels of LOTUSDIS. Our models degrade under severe reverberation and distant microphones (73--84\% CER on the Bluetooth microphones at 3\,m and 10\,m, which we exclude), leaving far-field Thai ASR outside the scope of this work. Finally, because Thai datasets and ASR models follow different writing conventions for numbers, loanwords, and colloquial particles, a standardized normalization pipeline is needed for fair comparison.

\section{Conclusion}
\label{sec:conclusion}
We present \hyb, a 33M-parameter FastConformer transducer with an auxiliary CTC head, together with \hybstream, \sysname, and a Rust inference engine. Despite having only 33M parameters, \hyb and \hybstream provide CER comparable to 115M- to 0.6B-parameter open-weight models on general Thai benchmarks. At decode time, \sysname spots listed keywords on the CTC posteriors, verifies each candidate under both the transducer and CTC heads, and rewrites only the verified spans in the transcript, raising entity recall on \cab from 32.7\% to 87.8\%. Paired with the Rust inference engine, the recognition and keyword-boosting pipeline runs with high throughput and low latency on x86-64 CPUs with Intel AMX, with support for other CPU architectures. We publicly release the model weights, \sysname, and the Rust inference engine.

\section*{Acknowledgments}
We thank Kunat Pipatanakul and Potsawee Manakul for their valuable support, feedback, and insights throughout this work. We also thank Wayu Research for \cab, NECTEC for the LOTUSDIS corpus and Pathumma models, the creators of the openly licensed Thai speech corpora used in training, and the NVIDIA NeMo team for their open-source toolkit. Finally, we are deeply grateful to the members of the Typhoon team, past and present. Their ideas, their work, and their shared commitment to Thai language technology made this work possible, and it belongs to them as much as to its authors.

\bibliographystyle{unsrtnat}
\bibliography{refs}

\begin{thebibliography}{36}
\providecommand{\natexlab}[1]{#1}
\providecommand{\url}[1]{\texttt{#1}}
\expandafter\ifx\csname urlstyle\endcsname\relax
  \providecommand{\doi}[1]{doi: #1}\else
  \providecommand{\doi}{doi: \begingroup \urlstyle{rm}\Url}\fi

\bibitem[Radford et~al.(2023)Radford, Kim, Xu, Brockman, Mcleavey, and Sutskever]{radford2023robust}
Alec Radford, Jong~Wook Kim, Tao Xu, Greg Brockman, Christine Mcleavey, and Ilya Sutskever.
\newblock Robust speech recognition via large-scale weak supervision.
\newblock In Andreas Krause, Emma Brunskill, Kyunghyun Cho, Barbara Engelhardt, Sivan Sabato, and Jonathan Scarlett, editors, \emph{Proceedings of the 40th International Conference on Machine Learning}, volume 202 of \emph{Proceedings of Machine Learning Research}, pages 28492--28518. PMLR, 23--29 Jul 2023.
\newblock URL \url{https://proceedings.mlr.press/v202/radford23a.html}.

\bibitem[Tipaksorn et~al.(2024)Tipaksorn, Sommuang, Chatthong, and Thangthai]{tipaksorn2024PathummaWhisper}
Pattara Tipaksorn, Wayupuk Sommuang, Oatsada Chatthong, and Kwanchiva Thangthai.
\newblock {Pathumma Whisper Large V3 (TH)}, 2024.
\newblock URL \url{https://huggingface.co/nectec/Pathumma-whisper-th-large-v3}.

\bibitem[Aung et~al.(2024)Aung, Thavornmongkol, Boribalburephan, Tangsriworakan, Pipatsrisawat, and Achakulvisut]{aung-etal-2024-thonburian}
Zaw~Htet Aung, Thanachot Thavornmongkol, Atirut Boribalburephan, Vittavas Tangsriworakan, Knot Pipatsrisawat, and Titipat Achakulvisut.
\newblock {Thonburian Whisper}: Robust fine-tuned and distilled whisper for {T}hai.
\newblock In Mourad Abbas and Abed~Alhakim Freihat, editors, \emph{Proceedings of the 7th International Conference on Natural Language and Speech Processing (ICNLSP 2024)}, pages 149--156, Trento, October 2024. Association for Computational Linguistics.
\newblock URL \url{https://aclanthology.org/2024.icnlsp-1.17}.

\bibitem[{Typhoon Team, SCB 10X}(2026{\natexlab{a}})]{typhoon_whisper_large}
{Typhoon Team, SCB 10X}.
\newblock {Typhoon-Whisper-Large-v3: Robust Thai Automatic Speech Recognition}.
\newblock \url{https://huggingface.co/typhoon-ai/typhoon-whisper-large-v3}, 2026{\natexlab{a}}.

\bibitem[Sirichotedumrong et~al.(2026{\natexlab{a}})Sirichotedumrong, Na-Thalang, Manakul, Taveekitworachai, Sripaisarnmongkol, and Pipatanakul]{typhoonrt}
Warit Sirichotedumrong, Adisai Na-Thalang, Potsawee Manakul, Pittawat Taveekitworachai, Sittipong Sripaisarnmongkol, and Kunat Pipatanakul.
\newblock Typhoon {ASR} real-time: {FastConformer}-transducer for {Thai} automatic speech recognition, 2026{\natexlab{a}}.
\newblock arXiv:2601.13044.

\bibitem[Sirichotedumrong et~al.(2026{\natexlab{b}})Sirichotedumrong, Samutsin, Wani, Sripaisarnmongkol, and Pipatanakul]{typhoonstreaming}
Warit Sirichotedumrong, Tanawin Samutsin, Shah~Faisal Wani, Sittipong Sripaisarnmongkol, and Kunat Pipatanakul.
\newblock Typhoon {ASR} streaming: Steerable low-latency {Thai} speech recognition with real-time shallow fusion, 2026{\natexlab{b}}.
\newblock arXiv:2609.14991.

\bibitem[{iApp Technology}(2026)]{openthai2asrstreaming}
{iApp Technology}.
\newblock {OpenThai2.0-ASR-Streaming}: streaming {Thai} speech recognition.
\newblock Hugging Face model card, 2026.
\newblock \url{https://huggingface.co/iapp/OpenThai2.0-ASR-Streaming}.

\bibitem[Shi et~al.(2026)Shi, Wang, Guo, Wang, Zhang, Zhang, Guo, Hao, Xi, Yang, Xu, Zhou, and Lin]{qwen3asr}
Xian Shi, Xiong Wang, Zhifang Guo, Yongqi Wang, Pei Zhang, Xinyu Zhang, Zishan Guo, Hongkun Hao, Yu~Xi, Baosong Yang, Jin Xu, Jingren Zhou, and Junyang Lin.
\newblock Qwen3-{ASR} technical report.
\newblock \emph{arXiv preprint arXiv:2601.21337}, 2026.

\bibitem[{Reson8}(2026)]{rumble2flash}
{Reson8}.
\newblock {Rumble-2-Flash}.
\newblock Hugging Face model card, 2026.
\newblock \url{https://huggingface.co/Reson8/Rumble-2-Flash}.

\bibitem[{NVIDIA}(2026{\natexlab{a}})]{nemotron35transcribe}
{NVIDIA}.
\newblock {NVIDIA-Nemotron-3.5-Transcribe-30B-A4B}.
\newblock Hugging Face model card, 2026{\natexlab{a}}.
\newblock \url{https://huggingface.co/nvidia/NVIDIA-Nemotron-3.5-Transcribe-30B-A4B}.

\bibitem[{Typhoon Team, SCB 10X}(2026{\natexlab{b}})]{typhoon_whisper_turbo}
{Typhoon Team, SCB 10X}.
\newblock {Typhoon-Whisper-Turbo: Efficient Thai Automatic Speech Recognition}.
\newblock \url{https://huggingface.co/typhoon-ai/typhoon-whisper-turbo}, 2026{\natexlab{b}}.

\bibitem[{NECTEC}(2026)]{pathummaRNNT}
{NECTEC}.
\newblock {Pathumma-stt-th-fastconformer-rnnt-large}: a {FastConformer} transducer for {Thai}.
\newblock \url{https://huggingface.co/nectec/Pathumma-stt-th-fastconformer-rnnt-large}, 2026.
\newblock FastConformer-Transducer, CC-BY-4.0.

\bibitem[Zhao et~al.(2019)Zhao, Sainath, Rybach, Rondon, Bhatia, Li, and Pang]{zhao2019shallow}
Ding Zhao, Tara~N. Sainath, David Rybach, Pat Rondon, Deepti Bhatia, Bo~Li, and Ruoming Pang.
\newblock Shallow-fusion end-to-end contextual biasing.
\newblock In \emph{Proc. Interspeech}, pages 1418--1422, 2019.
\newblock \doi{10.21437/Interspeech.2019-1209}.

\bibitem[Andrusenko et~al.(2025)Andrusenko, Bataev, Grigoryan, Lavrukhin, and Ginsburg]{andrusenko2025turbobias}
Andrei Andrusenko, Vladimir Bataev, Lilit Grigoryan, Vitaly Lavrukhin, and Boris Ginsburg.
\newblock {TurboBias}: Universal {ASR} context-biasing powered by {GPU}-accelerated phrase-boosting tree, 2025.
\newblock arXiv:2508.07014.

\bibitem[Andrusenko et~al.(2024)Andrusenko, Laptev, Bataev, Lavrukhin, and Ginsburg]{andrusenko2024fast}
Andrei Andrusenko, Aleksandr Laptev, Vladimir Bataev, Vitaly Lavrukhin, and Boris Ginsburg.
\newblock Fast context-biasing for {CTC} and transducer {ASR} models with {CTC}-based word spotter.
\newblock In \emph{Proc. Interspeech}, 2024.
\newblock arXiv:2406.07096.

\bibitem[Tsai et~al.(2026)Tsai, Lo, Sun, and Chen]{tsai2026streamingctcws}
Kai-Chen Tsai, Tien-Hong Lo, Yun-Ting Sun, and Berlin Chen.
\newblock Contextual biasing for streaming {ASR} via {CTC}-based word spotting, 2026.
\newblock arXiv:2605.18222.

\bibitem[Noroozi et~al.(2024)Noroozi, Majumdar, Kumar, Balam, and Ginsburg]{noroozi2024stateful}
Vahid Noroozi, Somshubra Majumdar, Ankur Kumar, Jagadeesh Balam, and Boris Ginsburg.
\newblock Stateful conformer with cache-based inference for streaming automatic speech recognition.
\newblock In \emph{Proc. ICASSP}, pages 12041--12045, 2024.
\newblock \doi{10.1109/ICASSP48485.2024.10446861}.

\bibitem[{NVIDIA}(2026{\natexlab{b}})]{nemotron35asrstreaming}
{NVIDIA}.
\newblock {nemotron-3.5-asr-streaming-0.6b}.
\newblock Hugging Face model card, 2026{\natexlab{b}}.
\newblock \url{https://huggingface.co/nvidia/nemotron-3.5-asr-streaming-0.6b}.

\bibitem[Rekesh et~al.(2023)Rekesh, Koluguri, Kriman, Majumdar, Noroozi, Huang, Hrinchuk, Puvvada, Kumar, Balam, and Ginsburg]{rekesh2023fast}
Dima Rekesh, Nithin~Rao Koluguri, Samuel Kriman, Somshubra Majumdar, Vahid Noroozi, He~Huang, Oleksii Hrinchuk, Krishna Puvvada, Ankur Kumar, Jagadeesh Balam, and Boris Ginsburg.
\newblock {Fast Conformer} with linearly scalable attention for efficient speech recognition.
\newblock In \emph{2023 IEEE Automatic Speech Recognition and Understanding Workshop (ASRU)}, pages 1--8, 2023.
\newblock \doi{10.1109/ASRU57964.2023.10389701}.

\bibitem[Xu et~al.(2023)Xu, Jia, Majumdar, Huang, Watanabe, and Ginsburg]{xu2023tdt}
Hainan Xu, Fei Jia, Somshubra Majumdar, He~Huang, Shinji Watanabe, and Boris Ginsburg.
\newblock Efficient sequence transduction by jointly predicting tokens and durations.
\newblock In \emph{Proceedings of the 40th International Conference on Machine Learning}, volume 202 of \emph{Proceedings of Machine Learning Research}, pages 38462--38484. PMLR, 2023.
\newblock URL \url{https://proceedings.mlr.press/v202/xu23g.html}.

\bibitem[Graves et~al.(2006)Graves, Fern{\'a}ndez, Gomez, and Schmidhuber]{graves2006connectionist}
Alex Graves, Santiago Fern{\'a}ndez, Faustino Gomez, and J{\"u}rgen Schmidhuber.
\newblock Connectionist temporal classification: Labelling unsegmented sequence data with recurrent neural networks.
\newblock In \emph{Proc. ICML}, 2006.

\bibitem[Kuchaiev et~al.(2019)Kuchaiev, Li, Nguyen, Hrinchuk, Leary, Ginsburg, Kriman, Beliaev, Lavrukhin, Cook, Castonguay, Popova, Huang, and Cohen]{kuchaiev2019nemo}
Oleksii Kuchaiev, Jason Li, Huyen Nguyen, Oleksii Hrinchuk, Ryan Leary, Boris Ginsburg, Samuel Kriman, Stanislav Beliaev, Vitaly Lavrukhin, Jack Cook, Patrice Castonguay, Mariya Popova, Jocelyn Huang, and Jonathan~M. Cohen.
\newblock {NeMo}: a toolkit for building {AI} applications using neural modules, 2019.
\newblock URL \url{https://arxiv.org/abs/1909.09577}.

\bibitem[Li et~al.(2023)Li, Takamichi, Saeki, Chen, Shiota, and Watanabe]{li2023yodas}
Xinjian Li, Shinnosuke Takamichi, Takaaki Saeki, William Chen, Sayaka Shiota, and Shinji Watanabe.
\newblock {YODAS}: {YouTube}-oriented dataset for audio and speech.
\newblock In \emph{IEEE Automatic Speech Recognition and Understanding Workshop (ASRU)}, 2023.

\bibitem[Chen et~al.(2026)Chen, Takamichi, Shiota, Fukayama, Cornell, and Watanabe]{chen2026yodas3}
William Chen, Shinnosuke Takamichi, Sayaka Shiota, Satoru Fukayama, Samuele Cornell, and Shinji Watanabe.
\newblock {YODAS} v3: Over 1 million hours of high-bandwidth, stereophonic, multilingual speech, 2026.
\newblock arXiv:2609.29448.

\bibitem[Suwanbandit et~al.(2023)Suwanbandit, Naowarat, Sangpetch, and Chuangsuwanich]{suwanbandit23_interspeech}
Artit Suwanbandit, Burin Naowarat, Orathai Sangpetch, and Ekapol Chuangsuwanich.
\newblock {Thai Dialect Corpus and Transfer-based Curriculum Learning Investigation for Dialect Automatic Speech Recognition}.
\newblock In \emph{Proc. INTERSPEECH 2023}, pages 4069--4073, 2023.
\newblock \doi{10.21437/Interspeech.2023-1828}.

\bibitem[Ardila et~al.(2020)Ardila, Branson, Davis, Kohler, Meyer, Henretty, Morais, Saunders, Tyers, and Weber]{ardila2020common}
Rosana Ardila, Megan Branson, Kelly Davis, Michael Kohler, Josh Meyer, Michael Henretty, Reuben Morais, Lindsay Saunders, Francis Tyers, and Gregor Weber.
\newblock Common voice: A massively-multilingual speech corpus.
\newblock In \emph{Proceedings of the 12th Language Resources and Evaluation Conference}, pages 4218--4222, 2020.

\bibitem[Tipaksorn et~al.(2026)Tipaksorn, Thatphithakkul, Chunwijitra, and Thangthai]{tipaksorn2025lotusdis}
Pattara Tipaksorn, Sumonmas Thatphithakkul, Vataya Chunwijitra, and Kwanchiva Thangthai.
\newblock {LOTUSDIS}: A thai far-field meeting corpus for robust conversational {ASR}.
\newblock In \emph{Proc. ICASSP}, 2026.
\newblock arXiv:2509.18722.

\bibitem[Wongpithayadisai et~al.(2026)Wongpithayadisai, Chaksangchaichot, Sangnark, Prakrankamanant, Gangwanpongpun, Boonpunmongkol, Milindasuta, Na-Pombejra, Nutanong, and Chuangsuwanich]{wongpithayadisai2026thaiser}
Jilamika Wongpithayadisai, Chompakorn Chaksangchaichot, Soravitt Sangnark, Patawee Prakrankamanant, Krit Gangwanpongpun, Siwa Boonpunmongkol, Premmarin Milindasuta, Dangkamon Na-Pombejra, Sarana Nutanong, and Ekapol Chuangsuwanich.
\newblock {Thai} speech emotion recognition ({THAI-SER}) corpus.
\newblock \emph{Language Resources and Evaluation}, 60:\penalty0 59, 2026.
\newblock \doi{10.1007/s10579-026-09932-w}.

\bibitem[Na-Thalang et~al.(2025)Na-Thalang, Wittayasakpan, Phatcharoen, and Buakaw]{nathalang2025developingopenconversationalspeech}
Adisai Na-Thalang, Chanakan Wittayasakpan, Kritsadha Phatcharoen, and Supakit Buakaw.
\newblock Developing an open conversational speech corpus for the isan language, 2025.
\newblock URL \url{https://arxiv.org/abs/2511.21229}.

\bibitem[Conneau et~al.(2023)Conneau, Ma, Khan, Zhang, Bapna, Boito, et~al.]{conneau2023fleurs}
Alexis Conneau, Min Ma, Simran Khan, Yu~Zhang, Ankur Bapna, Marcely~Zanon Boito, et~al.
\newblock {FLEURS}: Few-shot learning evaluation of universal representations of speech.
\newblock In \emph{2023 IEEE Spoken Language Technology Workshop (SLT)}, pages 798--805, 2023.

\bibitem[Yu et~al.(2021)Yu, Han, Gulati, Chiu, Li, Sainath, Wu, and Pang]{yu2021dualmode}
Jiahui Yu, Wei Han, Anmol Gulati, Chung-Cheng Chiu, Bo~Li, Tara~N. Sainath, Yonghui Wu, and Ruoming Pang.
\newblock Dual-mode {ASR}: Unify and improve streaming {ASR} with full-context modeling.
\newblock In \emph{International Conference on Learning Representations (ICLR)}, 2021.

\bibitem[Friedman(2001)]{friedman2001gbm}
Jerome~H. Friedman.
\newblock Greedy function approximation: A gradient boosting machine.
\newblock \emph{The Annals of Statistics}, 29\penalty0 (5):\penalty0 1189--1232, 2001.
\newblock \doi{10.1214/aos/1013203451}.

\bibitem[Pedregosa et~al.(2011)Pedregosa, Varoquaux, Gramfort, Michel, Thirion, Grisel, Blondel, Prettenhofer, Weiss, Dubourg, Vanderplas, Passos, Cournapeau, Brucher, Perrot, and Duchesnay]{pedregosa2011scikit}
Fabian Pedregosa, Ga{\"e}l Varoquaux, Alexandre Gramfort, Vincent Michel, Bertrand Thirion, Olivier Grisel, Mathieu Blondel, Peter Prettenhofer, Ron Weiss, Vincent Dubourg, Jake Vanderplas, Alexandre Passos, David Cournapeau, Matthieu Brucher, Matthieu Perrot, and {\'E}douard Duchesnay.
\newblock Scikit-learn: Machine learning in {Python}.
\newblock \emph{Journal of Machine Learning Research}, 12:\penalty0 2825--2830, 2011.

\bibitem[{Typhoon Team, SCB 10X}(2026{\natexlab{c}})]{typhoon_gigaspeech_test}
{Typhoon Team, SCB 10X}.
\newblock {Gigaspeech2-Typhoon: A Standardized Thai ASR Benchmark}.
\newblock \url{https://huggingface.co/datasets/typhoon-ai/gigaspeech2-typhoon}, 2026{\natexlab{c}}.

\bibitem[{Typhoon Team, SCB 10X}(2026{\natexlab{d}})]{tvspeech_dataset}
{Typhoon Team, SCB 10X}.
\newblock {TVSpeech: A Robust Thai ASR Test Set}.
\newblock \url{https://huggingface.co/datasets/typhoon-ai/TVSpeech}, 2026{\natexlab{d}}.

\bibitem[{Wayu Research}(2026)]{contextasrbench}
{Wayu Research}.
\newblock {thai-contextasr-bench}: a contextual-biasing benchmark for {Thai} {ASR}.
\newblock \url{https://huggingface.co/datasets/wayu-ai/thai-contextasr-bench}, 2026.
\newblock Dataset, lists and frozen scorer.

\end{thebibliography}

\appendix
\section{Verifier Features and Parameters}
\label{app:features}
\textbf{Search parameters.} During the posterior search, an alignment path can stay on the same vocabulary piece for up to 6 frames at a per-frame score of $\max(\text{blank},\text{repeat})$ and can skip at most one non-initial character with a penalty of $-5$. For a keyword of length $K$ characters, a candidate match proceeds to verification if its average score per character satisfies $s \ge -3$ and its lowest single-piece score satisfies $m_{\min} + 0.05K \ge -11$.

\textbf{Verifier training and hyperparameters.} We train the verification classifier on the internal banking benchmark using bias@50 and dis@50 (50 unspoken distractors only) keyword lists. The training set consists of all candidates that pass the search thresholds and modify the base text, totaling 2{,}753 candidates (496 positive) for \hyb and 4{,}589 candidates (517 positive) for \hybstream. The classifier is a gradient-boosted decision tree ensemble with 300 trees, a maximum of 15 leaves per tree, a learning rate of 0.05, and at least 40 samples per leaf. For dependency-free runtime inference, the trained trees are exported as plain arrays. We select the acceptance thresholds via out-of-fold cross-validation on the same internal benchmark by sweeping $\{0.4, 0.5, 0.6\}$ for Thai keywords and $\{0.2, 0.3, 0.4\}$ for Latin keywords to minimize the mean CER across bias@50 and dis@50. This yields thresholds of 0.6 (Thai) and 0.3 (Latin) for \hyb, and 0.6 (Thai) and 0.2 (Latin) for \hybstream. \Cref{tab:features} summarizes the 55 input features used by the classifier.

\begin{table}[!ht]\centering\small
\caption{Summary of the 55 verifier features. Here $s$ denotes the average search score per character, $m_{\min}$ the lowest single-piece score, $K$ the keyword length in characters, $d$ the character edit distance between the base text and the edited text, $\Delta_{\mathrm{TDT}}$ and $\Delta_{\mathrm{CTC}}$ the sequence log-probability differences under the transducer and CTC heads, and $L \in \{0,1\}$ the Latin-script indicator.}
\label{tab:features}
\setlength{\tabcolsep}{5pt}
\renewcommand{\arraystretch}{1.15}
\begin{tabular}{@{}lp{0.66\linewidth}@{}}
\toprule
\textbf{Feature Group} & \textbf{Included Features} \\
\midrule
Likelihood and edit (13) & Search scores ($s$ and $m_{\min} + 0.05K$), bounded keyword lengths ($\min(K,30)$ and $\max(0, 6-K)$), raw and normalized log-probability changes ($\Delta_{\mathrm{TDT}}$ and $\Delta_{\mathrm{CTC}}$ in total, per character $K$, and per edit distance $d$), whether the edit changes the base text, bounded edit distance $\min(d, 40)$, and relative edit distance $d/K$. \\
Script interaction (14) & Latin-script indicator $L$, plus the product of $L$ with each of the 13 likelihood and edit features above. \\
Cross-script similarity (8) & Character similarity between the replaced span and the keyword (both directly and against its best-matching phonetic alias across scripts, plus their difference), phonetic key similarity, script mismatch indicator, replaced span length relative to $K$, and exact presence of the keyword in the base text or greedy CTC text. \\
Span and list context (20) & Frame span length (total and per character), mean and minimum greedy CTC posterior within the span, blank-frame ratio, emitted subword pieces per character, number of competing candidates on overlapping frames, highest competing search score along with the candidate's margin and rank against it, total candidates from the list, keyword length (in characters, subword pieces, and words), fraction of digits and Latin characters, fuzzy match score in the base text and greedy CTC text, edit distance per character, and score spread $m_{\min} - s$. \\
\bottomrule
\end{tabular}
\end{table}
\newpage
\section{Training Hyperparameters}
\label{app:hparams}
\Cref{tab:hparams-hybrid} lists the full training hyperparameters for \hyb and \hybstream. Both models are trained with AdamW ($\beta{=}(0.9, 0.98)$, weight decay $10^{-3}$), gradient clipping at 1.0, a cosine learning-rate schedule decaying to $10^{-6}$, bfloat16 mixed precision, SpecAugment (2 frequency masks of width 27 and 10 time masks of width 5\%), FastEmit regularization ($\lambda{=}10^{-3}$), duration-bucketed batches of 200\,s of audio with gradient accumulation 2, and a fixed random seed (seed 0). Both share a 2{,}048-piece SentencePiece BPE tokenizer trained on the training labels with the full Thai character set, Latin letters (a to z, A to Z), and digits (0 to 9) included in the vocabulary.

\newcommand{\hptab}[2]{\begin{minipage}[t]{0.47\textwidth}\centering
{\scriptsize\textbf{#1}}\\[2pt]
{\scriptsize\setlength{\tabcolsep}{3pt}\renewcommand{\arraystretch}{1.05}
\begin{tabular}{@{}ll@{}}\toprule #2 \bottomrule\end{tabular}}\end{minipage}}

\begin{table}[!htb]\centering
\caption{\textbf{Training hyperparameters of \hyb and \hybstream.}}\label{tab:hparams-hybrid}
\noindent\hbox to\linewidth{%
\hptab{\hyb}{
Architecture & FastConformer, 16 layers, $d{=}256$, 4 heads \\
Heads & TDT (durations 0--4) + CTC (weight 0.3) \\
Initialization & random \\
Trainable parameters & 33.1M (all) \\
Output vocabulary & BPE, 2{,}048 pieces \\
Attention context & whole utterance \\
Conv.\ / subsampling & non-causal / non-causal ($8{\times}$, 80\,ms) \\
Peak learning rate & $3{\times}10^{-4}$ \\
Schedule & cosine, 5{,}000-step warmup, min $10^{-6}$ \\
Batching & 200\,s audio, grad.\ accum.\ 2 \\
Utterance length & 0.3--30\,s, no concatenation \\
Steps (10 epochs) & 379{,}747 \\
Training data & 4{,}219.4\,h Thai (\Cref{tab:data}) \\
Precision / peak mem.\ & bf16 / 13.6\,GB \\
Hardware / wall time & 1$\times$H100 / 15.9\,h \\}%
\hfil
\hptab{\hybstream}{
Attention contexts & [130,\,12], [70,\,6], [$-1$,\,$-1$] \\
Conv.\ / subsampling & causal / causal ($8{\times}$, 80\,ms) \\
Initialization & \hyb \\
Trainable parameters & 33.1M (all) \\
Output vocabulary & BPE, 2{,}048 pieces \\
Peak learning rate & $1.5{\times}10^{-4}$ \\
Schedule & cosine, 1{,}000-step warmup, min $10^{-6}$ \\
Batching & 200\,s audio, grad.\ accum.\ 2 \\
Concatenation & $p{=}0.7$, factor 6, max 40\,s \\
Distillation & dual-mode, KL on CTC (weight 1.0) \\
Steps (8 epochs) & 303{,}797 \\
Training data & 4{,}219.4\,h Thai (\Cref{tab:data}) \\
Precision / peak mem.\ & bf16 / 34.8\,GB \\
Hardware / wall time & 1$\times$H100 / 25.3\,h \\}%
}
\end{table}

\end{document}